\pdfoutput=1

\documentclass[11pt]{article}

\usepackage[final]{acl}
\usepackage{amsmath}
\usepackage{times}
\usepackage{latexsym}
\usepackage{graphicx}
\usepackage{booktabs}
\usepackage{multirow}
\usepackage[T1]{fontenc}

\usepackage[utf8]{inputenc}
\usepackage{mathrsfs}  
\usepackage{microtype}
\usepackage{enumitem}

\usepackage{amsfonts}

\title{Enhancing Advanced Visual Reasoning Ability of Large Language Models}

\author{
 \textbf{Zhiyuan Li\textsuperscript{}},
 \textbf{Dongnan Liu\textsuperscript{}},
 \textbf{Chaoyi Zhang\textsuperscript{}},
\\
 \textbf{Heng Wang\textsuperscript{}},
 \textbf{Tengfei Xue\textsuperscript{}},
 \textbf{Weidong Cai\textsuperscript{}}
\\
 \textsuperscript{}School of Computer Science, The University of Sydney
\\
 \small{\{zhli0736, czha5168, hwan9147, txue4133\}@uni.sydney.edu.au 
 }
 \\
 \small{
 \{dongnan.liu, tom.cai\}@sydney.edu.au
 }
}

\begin{document}
\maketitle
\begin{abstract}
Recent advancements in Vision-Language (VL) research have sparked new benchmarks for complex visual reasoning, challenging models' advanced reasoning ability. Traditional Vision-Language Models (VLMs) perform well in visual perception tasks while struggling with complex reasoning scenarios. Conversely, Large Language Models (LLMs) demonstrate robust text reasoning capabilities; however, they lack visual acuity. To bridge this gap, we propose \textbf{C}omplex \textbf{V}isual \textbf{R}easoning \textbf{L}arge \textbf{L}anguage \textbf{M}odels (CVR-LLM), capitalizing on VLMs' visual perception proficiency and LLMs' extensive reasoning capability. Unlike recent multimodal large language models (MLLMs) that require a projection layer, our approach transforms images into detailed, context-aware descriptions using an iterative self-refinement loop and leverages LLMs' text knowledge for accurate predictions without extra training. We also introduce a novel multi-modal in-context learning (ICL) methodology to enhance LLMs' contextual understanding and reasoning. Additionally, we introduce Chain-of-Comparison (CoC), a step-by-step comparison technique enabling contrasting various aspects of predictions. Our CVR-LLM presents the first comprehensive study across a wide array of complex visual reasoning tasks and achieves SOTA performance among all.
\end{abstract}

\section{Introduction}
\label{intro}
\begin{figure*}[t]
\vspace{-0.4cm}
  \centering
  \scalebox{0.95}{
  \includegraphics[width=\textwidth]{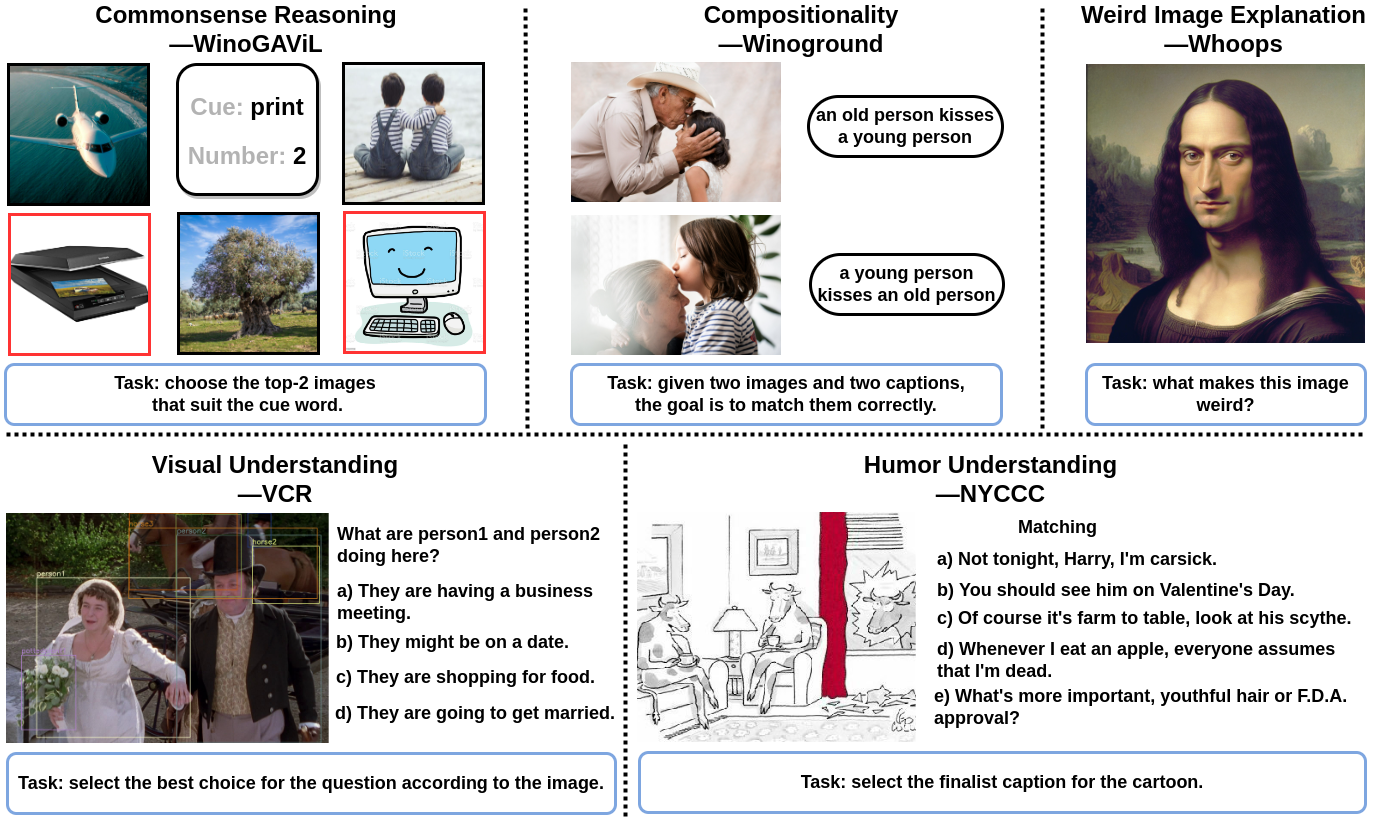}}
  \vspace{-0.2cm}
  \caption{Five distinct examples from diverse datasets in the complex visual reasoning field~\cite{bitton2023breaking} challenge AI models' ability of complex reasoning in different aspects such as general commonsense.}
  \vspace{-0.4cm}
  \label{fig1}
\end{figure*}

The concept of complex visual reasoning was introduced with Visual Commonsense Reasoning (VCR) dataset~\cite{zellers2019recognition} in 2019, which tests models' ability to understand visual content as well as commonsense cognition. However, the development in this field has remained relatively subdued, primarily due to Vision-Language Models' (VLMs) limitations in incorporating commonsense knowledge~\cite{gan2022vision}. Recent years have seen significant advancements in complex linguistic reasoning tasks~\cite{cobbe2021training,wei2022chain} due to the emerging GPT3~\cite{brown2020language}, LLaMA~\cite{touvron2023llama1}, and Vicuna~\cite{chiang2023vicuna}. This leap forward has triggered a renewed interest in the complex visual reasoning area, exploring how visual perception can enhance linguistic inference and potentially overcome previous hurdles~\cite{gan2022vision}. It has led to innovative benchmarks focusing on various aspects: commonsense reasoning - WinoGAViL~\cite{bitton2022winogavil}, compositionality - Winoground~\cite{thrush2022winoground}, weird image explanation - Whoops~\cite{bitton2023breaking}, and humor understanding - NYCCC~\cite{hessel2022androids}. These tasks demand models not only accurately interpret image content, but also integrate knowledge from daily experiences, general commonsense, cultural context, and humor sense. For example, a synthetic image, as shown in Whoop's example in Figure~\ref{fig1} of ``The portrait of the Mona Lisa depicts a stern male face.'' contradicts the cultural context, as the famous painting Mona Lisa depicts a female face. 

\begin{figure*}[t]
\vspace{-0.4cm}
   \includegraphics[width=0.95\linewidth]{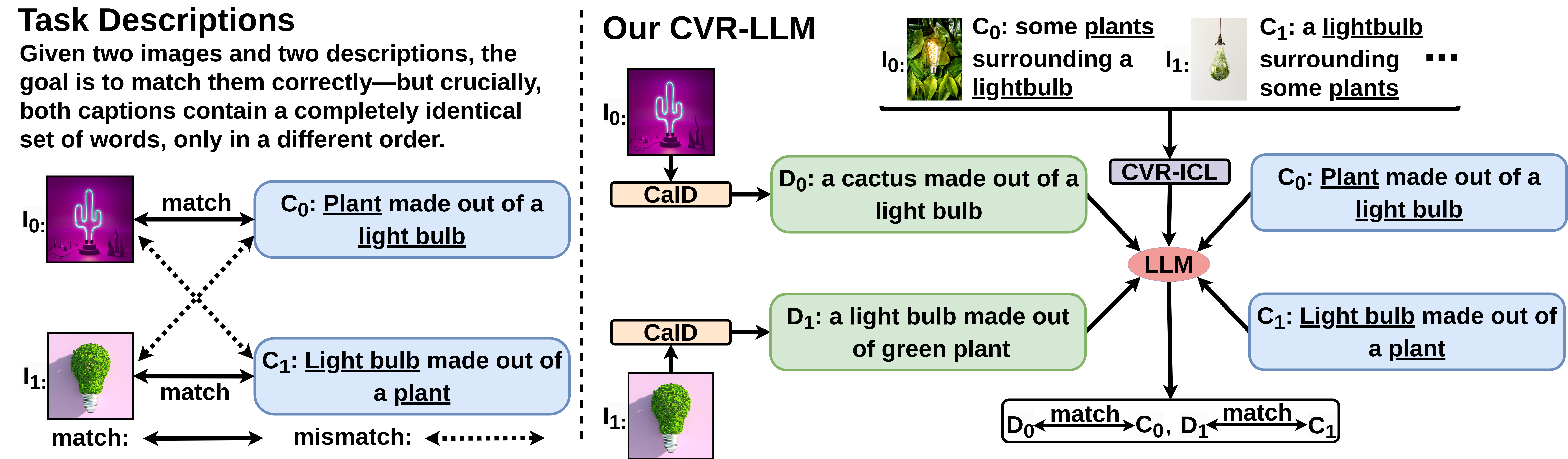}
    \vspace{-0.2cm}
   \caption{An example of our CVR-LLM works on the Winoground dataset. Our method transfers images into context-aware image descriptions through CaID and leverages the sophisticated reasoning and ICL abilities of LLMs with the CVR-ICL module, offering a more precise answer.}
   \label{fig2}
    \vspace{-0.4cm}
\end{figure*}

In this paper, we introduce a novel method named Complex Visual Reasoning Large Language Models (CVR-LLM), based on the "VLMs + LLMs" concept. Recent multimodal large language models (MLLMs) like LLaVA~\cite{liu2024visual, liu2023improved} and MiniGPT4~\cite{zhu2023minigpt,chen2023minigpt} have proven effective in many VL tasks. However, these models are resource-intensive, relying on millions of image-text pairs for projection layer learning. To overcome this limitation, our approach leverages the visual perception strengths of VLMs to translate images into context-aware image descriptions (CaID) via an inference-only, dual-loop self-refinement process that incorporates feedback from LLMs. These detailed descriptions enhance the LLMs' inference process, transforming multi-modal tasks into simpler single-modal challenges and streamlining the overall process. In addition, we develop a unique multi-modal in-context learning (ICL) approach named Complex Visual Reasoning ICL (CVR-ICL), which enhances the reasoning capacities of LLMs within a range of complex multi-modal environments. Figure~\ref{fig2} provides an illustration of how our CVR-LLM is applied to the Winoground task. It describes the images as appropriate sentences via CaID and utilizes the sophisticated reasoning and ICL abilities of LLMs through CVR-ICL for more accurate predictions. 

Our research stands as the pioneering study to explore such a broad array of benchmarks (WinoGAViL, Winoground, Whoops, VCR, and NYCCC), proposing a paradigm centred on the "VLM+LLM" concept for addressing complex visual reasoning tasks. Experimental results show that CVR-LLM achieves SOTA performance across all five tasks. Further ablation studies and comparative analyses reveal the effectiveness of each module and the superiority of our method over previous approaches. Particularly in comparative analysis, we introduce the Chain-of-Comparison (CoC) technique, inspired by "Chain-of-Thought" and utilizing GPT4~\cite{achiam2023gpt}, to address the limitations of conventional metrics in evaluating abstract concepts. CoC provides a nuanced analysis by systematically dissecting and quantitatively contrasting various facets of the results for a comprehensive evaluation.

Our contributions are summarized as follows: (1) We present the first comprehensive study across all complex visual reasoning tasks, including WinoGAViL, Winoground, Whoops, VCR, and NYCCC. (2) We design a context-aware image description generation method and a specific in-context learning strategy\footnote{The project is available at: https://CVR-LLM.github.io}, to enhance the advanced visual reasoning ability of LLMs to multi-modal complex visual reasoning tasks. (3) We further introduce Chain-of-Comparsion, a novel GPT4-based comparison technique inspired by "Chain-of-Thought" filling the gaps of traditional metrics in abstract concept evaluation. (4) Experimental results show that our approach surpasses current SOTA models in a range of complex visual reasoning scenarios.

\section{Related Work}
\subsection{Reasoning Research in Vision-Language Domain}
In recent years, multi-modal reasoning research has significantly advanced. Beyond the complex visual reasoning benchmarks discussed in Section~\ref{intro}, many studies focus on the reasoning process itself, such as chain-of-thought~\cite{kojima2022large,shaikh2022second} or reasoning modules~\cite{zhou2023uncovering,jiang2023multi}, which are crucial for enhancing AI models' analytical capabilities and performance. For instance, \citet{liu2023matcr} introduced a modality-aligned thought chain reasoning framework to incorporate explicit reasoning into task-oriented dialogue generation, improving contextual understanding and effectiveness. \citet{lv2023counterfactual} proposed a counterfactual cross-modality reasoning method for better video moment localization. \citet{zhou2023learning} developed a multi-step reasoning probability transfer mechanism to improve multi-label interaction classifications. \citet{yu2023hierarchical} presented a hierarchical reasoning network to consolidate multi-level interactive cues, from coarse to fine-grained details, enhancing Human-Object Interaction (HOI) representations.

\subsection{Large Language Models for Vision-Language Analysis}
The past two years have seen an unprecedented surge in the development and application of LLMs~\cite{brown2020language,touvron2023llama1,chiang2023vicuna} across diverse fields. LLMs have garnered acclaim for their robust capabilities, including advanced analytical prowess~\cite{kojima2022large}, extensive text-level knowledge~\cite{naveed2023comprehensive} and superior understanding ability~\cite{chang2023survey}. Furthermore, they are equipped with two powerful mechanisms: chain-of-thought~\cite{kojima2022large} and in-context learning~\cite{liu2021makes}, which significantly augment their effectiveness and performance in specialized tasks~\cite{naveed2023comprehensive}. For example, \citet{muraoka2023cross} developed a cross-lingual model trained alongside a cross-lingual LLM, leveraging LLMs' capabilities across languages. \citet{lan2023improving} proposed reasoning question prompts for Visual Question Answering (VQA) tasks, unlocking LLMs' potential in zero-shot learning. Additionally, \citet{yang2023against} introduced SODA, a system that integrates LLMs with explainable AI to assist marketers with data interpretation, enhancing human-AI collaboration. \citet{zhong2023adapter} used knowledge distillation to imbue the SUR-adapter with LLMs' semantic understanding and reasoning capabilities.


\section{Methods}
In this section, we introduce the CVR-LLM framework, highlighting its innovative process for generating context-aware image descriptions (CaID) as well as its complex visual reasoning in-context learning (CVR-ICL) strategy. Initially, we explain the CaID generation process, which differs from traditional image captioning by using a self-refinement loop with feedback from Large Language Models (LLMs) to produce accurate and contextually relevant descriptions (Section~\ref{description}). Subsequently, we present the CVR-ICL approach (Section~\ref{in-context}), which enhances LLMs' contextual understanding and reasoning by assessing relevant cases and selecting suitable complex multi-modal demonstrations.

\subsection{Context-Aware Image Description}
\label{description}
Pre-trained VLMs~\cite{li2023blip,alayrac2022flamingo} have demonstrated their proficiency in generating detailed image captions on benchmarks such as MSCOCO~\cite{chen2015microsoft}. However, while these captions may accurately reflect visual content, they are not customized for complex visual reasoning scenarios. Recently, the trend of multi-modal instruction-following agents like miniGPT4~\cite{zhu2023minigpt, chen2023minigpt} and LLaVA~\cite{liu2024visual, liu2023improved}, integrating open-source LLMs~\cite{chiang2023vicuna, touvron2023llama2} with pre-trained vision encoders~\cite{dosovitskiy2020image, liu2021swin} to create a MLLM, has become very popular. The effectiveness of these models is heavily reliant on tuning with vast amounts of VL instruction data, which is generated by powerful LLMs like ChatGPT~\cite{2022Introducingchatgpt} and GPT4~\cite{achiam2023gpt}. While promising, their reliance on extensive VL instruction data for tuning requires the substantial resource and time investment. In this work, we introduce a more efficient method for generating context-aware image descriptions, which depends on the inference process and leverages task-specific information and feedback from LLMs to craft better prompts, guiding the caption generation process more effectively.

\begin{figure}[t]
  \centering
   \includegraphics[width=0.98\linewidth]{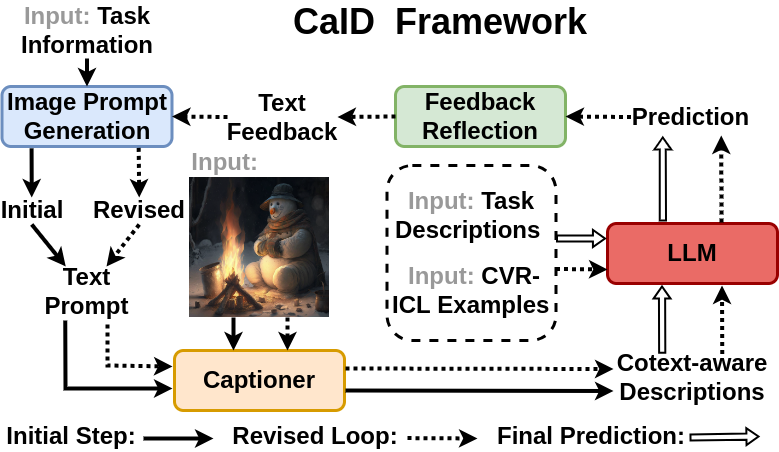}
   \vspace{-0.2cm}
   \caption{The framework overview of CaID. It is designed to transfer images into contextualized descriptions, bypassing the need for direct multi-modal fusion and leveraging LLMs' extensive knowledge for more accurate predictions.}
   \label{fig3}
   \vspace{-0.5cm}
\end{figure}

Our CaID framework optimizes the process of creating context-aware image descriptions through a dual-loop self-refinement approach, as shown in Figure~\ref{fig3}. Initially, it leverages task-specific details and LLM insights to craft precise image prompts. These initial prompts are designed to distill essential task-related information, guiding the captioner in producing descriptions that not only cover image content but are also deeply aligned with the task's requirements. Specifically, given a task specific text description $t$ with an image $i$ (for processes involving multiple images, we approach each image sequentially), the generation of initial context-aware image descriptions can be described as follows:
 \begin{equation}
\begin{aligned}
    d_{init} &= \textit{C} (i, \textit{L}(t)),
  \label{eq1}
 \end{aligned}
\end{equation}
where $d_{init}$ is the initial generated context-aware image description. $\textit{C}$ is the image-to-text captioner, transfering the image into the description. $\textit{L}$ is the LLM, encapsulating crucial task-related text information $t$ (e.g. requirements, questions, cue words) into feature prompts.

In the second loop, our approach is crafted to encapsulate essential task-related details as well as LLMs' feedback, enhancing description generation with LLMs' vast knowledge. Specifically, it merges initial descriptions with task specifics and CVR-ICL examples into a task-focused prompt, guiding LLMs to make more precise predictions. These predictions are then treated as pseudo labels, asking LLMs to design further inquiries for deeper insights around them. In this way, we build up a feedback reflection between LLM prediction and context-aware caption, enhancing the richness and accuracy of the content produced. The textual feedback is then leveraged to refine the image prompts, providing deep insights that inform and guide the generation of nuanced image descriptions. The revised context-aware image descriptions can be described as follows:
 \begin{equation}
\begin{aligned}
    d_{revised} &= \textit{C} (i, \textit{L}(t, \textit{Q}(p))),
  \label{eq2}
 \end{aligned}
\end{equation}
where $d_{revised}$ is the revised generated context-aware image description. $\textit{Q}$ is the further query from LLM. $p$ is the prediction from LLM according to the generated task prompt. $\textit{Q}(p)$ is the text feedback for updating image prompt.

\begin{figure*}[t]
\vspace{-0.4cm}
  \centering
   \includegraphics[width=0.89\linewidth]{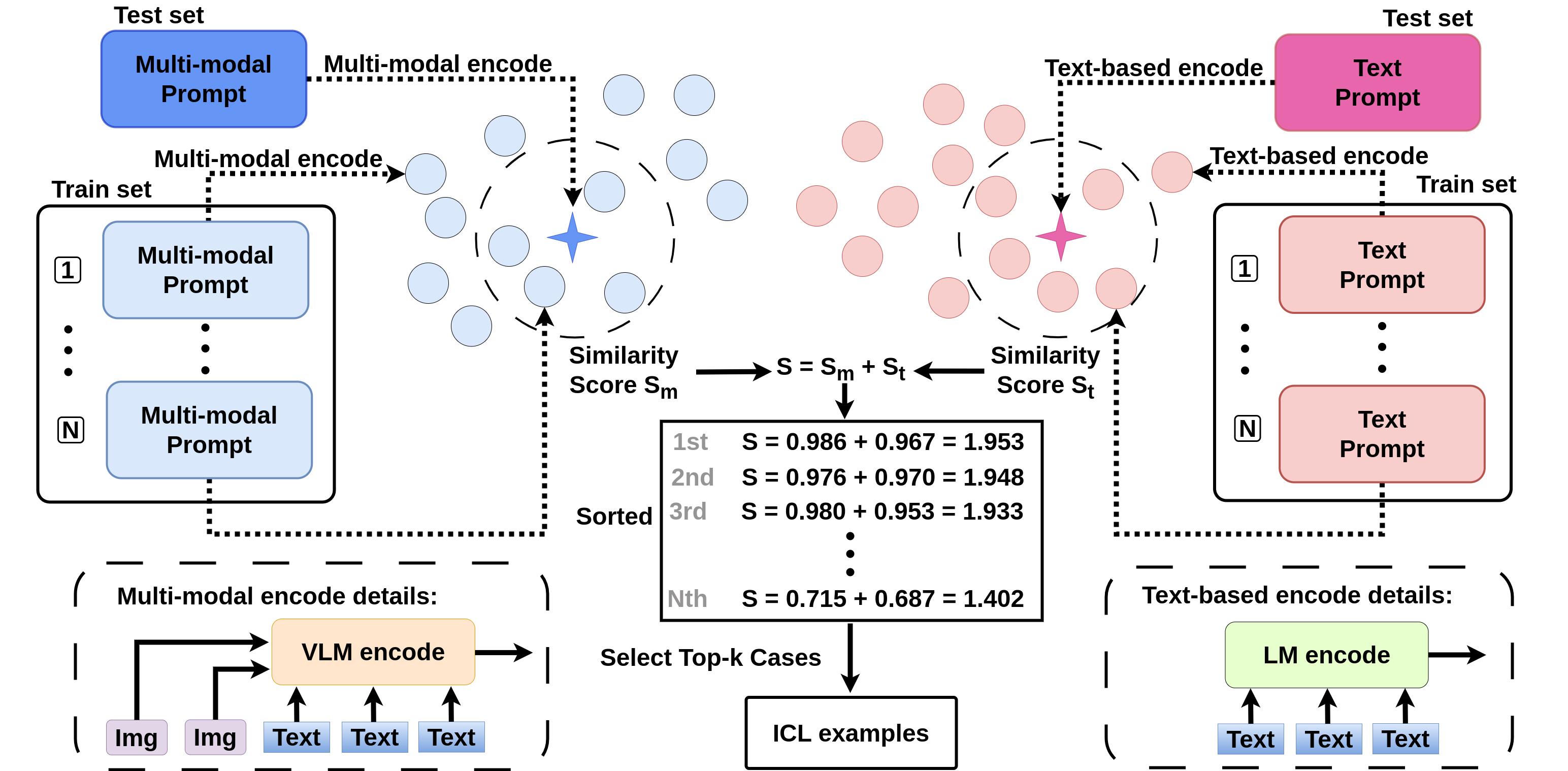}
   \vspace{-0.2cm}
   \caption{The generic diagram of our proposed CVR-ICL approach. The dual analysis enables our approach to more effectively select contextually relevant examples from text and multi-modal domains.}
   \label{fig4}
   \vspace{-0.4cm}
\end{figure*}

\subsection{Complex Visual Reasoning ICL}
\label{in-context}
LLMs are renowned for their exceptional in-context learning capabilities, especially with task-specific examples. The optimal in-context exemplars enable LLMs to leverage their background knowledge for more precise outcomes. However, most of the research works~\cite{liu2021makes,sorensen2022information} have primarily focused on the text-centric domain, with few works~\cite{alayrac2022flamingo,zhao2023mmicl} exploring multi-modal in-context learning for VL tasks. Our approach, unlike prior methods focused solely on text similarity in NLP, such as the $k$NN-augmented in-context example selection (KATE), integrates multi-modal factors, thereby enriching the discipline with a fresh perspective. Furthermore, it is also different from MMICL~\cite{zhao2023mmicl} in the multi-modal domain, which employs a vision prompt generator for image-to-visual embedding conversion and merges these with text embeddings as a union measurement factor. 

Complex visual reasoning tasks demand models capable of selecting in-context examples from a multi-modal domain, leveraging extensive background knowledge and information within it~\cite{zhao2023mmicl}. However, our CVR-LLM is grounded in LLMs, which are inherently text-based, leading to a gap between textual and multi-modal domains. Directly applying a text-based $k$NN clustering method could result in the loss of important multi-modal information. On the other hand, using multi-modal information for retrieval might ignore essential context-aware information within our generated image descriptions. To address this, we propose the complex visual reasoning ICL, which aims to select in-context examples for LLMs by effectively integrating both text and multi-modal components. This dual analysis enables our LLM to more effectively select contextually relevant examples, ensuring a balanced integration of text and multi-modal insights for enhanced in-context learning. Figure~\ref{fig4} illustrates the framework of our CVR-ICL strategy. Specifically, given a task $t$ with an image $i$, we initially convert the image into a description $d$, which enables the task to be applicable not only in multi-modal domains but also in text-only scenarios. Then, we employ a multi-modal encoder $\textit{f}_{m}$ and a text encoder $\textit{f}_{t}$ to transform inputs from the multi-modal domain and the text domain into vector representations as follows:
\begin{subequations}
\begin{align}
    &x_m= \textit{f}_{m}(t,i), \label{eq3-1}\\
    &x_t= \textit{f}_{t}(t,d), \label{eq3-2}
 \end{align}
\end{subequations}
where $x_m$ is the vector representation in the multi-modal domain. $x_t$ is the vector representation in the text domain.

Upon transforming each example into two distinct vector forms, we compute the cosine similarity score to identify and select the examples that are most relevant. Considering a target sample in test set and the $i$th example in the training set, the similarity calculation process can be expressed as follows: 
\begin{subequations}
\begin{align}
    &s_m= \textit{f}_{c}(x_m,x_m^{\textit{i}th}), \label{eq4-1}\\
    &s_t= \textit{f}_{c}(x_t,x_t^{\textit{i}th}),  \label{eq4-2}\\
    &s = s_m + s_t, \label{eq4-3}
\end{align}
\end{subequations}
where $s_m$ is the similarity score between the target sample and $i$th example in dataset on the multi-modal domain, $s_t$ is the similarity score between the target sample and $i$th example in dataset on the text domain. $s$ is the final similarity score. $\textit{f}_c$ is the cosine similarity function. Finally, the top-$k$ cases with the highest $s$ are selected as the in-context examples, aimed at boosting the contextual understanding and prediction accuracy of the LLMs.

\begin{table*}[t]
\centering
\vspace{-0.4cm}
 \scalebox{0.68}{
\begin{tabular}{lc|ccc|ccc|c|cc|cc}
\toprule[0.6mm]
\multirow{2}{*}{Type}                       & \multirow{2}{*}{Model} & \multicolumn{3}{c|}{WinoGAViL} & \multicolumn{3}{c|}{Winoground} & Whoops     & \multicolumn{2}{c|}{VCR}                                    & \multicolumn{2}{c}{NYCCC}        \\
                                            &                        & 5/6      & 10/12     & SWOW    & Text     & Image     & Group    & GPT4 Rate & Q-\textgreater{}A & QA-\textgreater{}R & Match acc. & CrowdAcc \\ 
                                            \midrule[0.3mm]
\multirow{6}{*}{VLM}                        & ViLT~\shortcite{kim2021vilt}                   & 55.0     & 52.0      & 59.0    & 34.7     & 14.0      & 9.2      & -          & -                 & -                  & -             & -      \\
                                            & CLIP ViT-L/14~\shortcite{radford2021learning}          & 47.0     & 15.0      & 66.0       & -         & -        & -          & -                 & -                  & -                  & 56.6          & 55.8     \\
                                            & UNITER~\shortcite{chen2020uniter}                 & -        & -         & -       & 38.0     & 14.0      & 10.5     & -          & -                 & -                  & -                  & -        \\
                                            & ViLLA~\shortcite{gan2020large}                  & -        & -         & -       & 37.0     & 13.2      & 11.0     & -          & -                 & -                             & 48.1          & 47.0     \\
                                             & BLIP~\shortcite{li2022blip}                   & 54.6     & 45.0      & 66.5    & \textbf{46.5}     & 27.7      & 24.2     & 22.0       & 29.2              & 27.5                       & 58.7          & 58.1    \\
                                            & BLIP2~\shortcite{li2023blip}                  & 49.3     & 38.8      & 71.6    & 44.0     & 26.0      & 23.5     & 31.0       & 24.5              & 25.6                      & 58.3          & 56.7     \\
                                            \midrule[0.3mm]
\multicolumn{1}{l}{\multirow{4}{*}{MLLM}} & LLaVA 1.0~\shortcite{liu2024visual}              & -        & -         & -       & -        & -         & -        & 32.0       & 28.3              & 40.0                         & 55.8          & 53.1    \\
\multicolumn{1}{c}{}                        & LLaVA 1.5~\shortcite{liu2023improved}              & -        & -         & -       & -        & -         & -        & 42.4       & 35.1              & 44.5                       & 59.3          & 56.0      \\
\multicolumn{1}{c}{}                        & MiniGPT4 V1~\shortcite{zhu2023minigpt}          & -        & -         & -       & -        & -         & -        & 44.6       & 40.6              & 47.7                           & 58.5          & 55.6       \\
\multicolumn{1}{c}{}                        & MiniGPT4 V2~\shortcite{chen2023minigpt}          & -        & -         & -       & -        & -         & -        & 48.2       & 48.8              & 49.7                           & 60.4          & \textbf{59.2}      \\
\midrule[0.3mm]
\multicolumn{1}{l}{\multirow{3}{*}{VLM+LLM}}                 & CVR-LLM$_{Llama3}$                & 72.3     & 70.4      & \textbf{88.7}    & 45.0     & 29.5      & 24.5     & 60.4      & 50.5        & 52.4                    & 59.8          & 57.7      \\
\multicolumn{1}{c}{}                         & CVR-LLM$_{GPT3.5}$                & 73.4     & 71.6      & 83.4    & 42.7     & 30.5      & 23.5     & 61.2      & 51.1        & 53.4                    & 59.4          & 56.8      \\
\multicolumn{1}{c}{}              & CVR-LLM$_{GPT4}$                & \textbf{74.7}     & \textbf{73.2}      & 86.5    & 43.5     & \textbf{35.0}      & \textbf{26.5}     & \textbf{62.0}      & \textbf{52.9}        & \textbf{54.3}                    & \textbf{60.6}          & 57.4      \\
\toprule[0.6mm]
\end{tabular}}
\vspace{-0.2cm}
\caption{The comparison of our CVR-LLM with popular VLMs and MM LLMs on five complex visual reasoning tasks. Notably, MLLMs like LLaVA and MiniGPT4 exhibit limitations in handling tasks involving multiple images or computing image-text similarity scores, resulting in their performance being unavailable for tasks like WinoGAViL and Winoground.}
\vspace{-0.4cm}
\label{tab1}
\end{table*}

\section{Experiments}
\subsection{Dataset and Metrics}
To evaluate the effectiveness of our proposed method, we conduct a comprehensive test in complex visual reasoning areas. Our evaluation included WinoGAViL (4373 samples), Winoground (400 samples), Whoops (500 samples), VCR (2653 out of over 26k samples, selecting a random 10$\%$), and NYCCC (528 samples), providing a broad assessment of our approach's capabilities.
In the terms of metrics, we adhered to the evaluation methods provided by these datasets, ensuring a fair assessment of our method's performance. 

\subsection{Implementation Details}
For the basic captioner in context-aware image description (Section~\ref{description}), we choose the BLIP2-flant5xxl~\cite{li2023blip} as our baseline. For CVR-ICL phase (Section~\ref{in-context}), we employ BM25~\cite{robertson1995okapi} and BLIP2 multi-embedding~\cite{li2023blip} to encode text and multi-modal inputs, respectively. It is important to note that the ICL example results are derived from LLM inference without using actual annotations to prevent data leakage. For our LLMs, we choose three popular LLMs as inference models for generation tests including: Llama3-8B~\cite{llama3modelcard} for CVR-LLM$_{Llama3}$, GPT3.5~\cite{openai2023gpt3.5} for CVR-LLM$_{GPT3.5}$, and GPT4~\cite{achiam2023gpt} for CVR-LLM$_{GPT4}$. Performance comparisons are conducted directly on the test set without any fine-tuning, as WinoGAViL, Winoground, and NYCC datasets are exclusively for testing purposes.

\subsection{Comparison to State-of-the-Arts}
In this section, we evaluate our proposed CVR-LLM against various models across a range of complex visual reasoning tasks, including WinoGAViL, Winoground, Whoops, VCR, and NYCCC. These models fall into two categories: VLMs~\cite{kim2021vilt, radford2021learning, gan2020large, li2023blip} and MLLMs~\cite{liu2024visual, liu2023improved, zhu2023minigpt, chen2023minigpt}. Notably, MLLMs like LLaVA and MiniGPT4 struggle with tasks involving multiple images, making their performance data unavailable for WinoGAViL and Winoground.

Table~\ref{tab1} showcases our method's superiority across five tasks, eclipsing both VLMs and LMMs. For example, our CVR-LLM$_{Llama3}$ significantly surpasses the SOTA model BLIP2 by achieving an $88.7\%$ accuracy (+17.1 improvement) in SWOW setting on the WinoGAViL benchmarks. Similarly, it outperforms the SOTA model MiniGPT4 with a $62.0\%$ accuracy (+13.8 improvement) on the GPT4 rate~\cite{bitton2023breaking} for Whoops tasks, underscoring our framework's advanced performance. Additionally, our method performs well on three LLM-based categories, demonstrating robust generation abilities with consistent performance. This highlights the versatility and adaptability of our model, ensuring high-quality results across various complex visual reasoning tasks.

\subsection{Ablation Studies}
In this section, we examine the individual contributions of the components within our framework CVR-LLM$_{GPT4}$. As demonstrated in Table~\ref{tab2}, we present an ablation study that quantifies the performance impact of each module across various datasets. The experimental findings suggest that the CVR-ICL module significantly boosts the inference performance of LLMs compared to using context-aware image descriptions alone, with the exception of the NYCCC dataset (It may be due to NYCCC's focus on humor, where precise descriptions are more critical). This highlights the CVR-ICL module's effectiveness in enhancing LLM capabilities across various tasks. In addition, our comprehensive method, CVR-LLM, which integrates both context-aware descriptions and CVR-ICL, achieves a substantial enhancement in performance relative to the baseline.

\begin{table*}[t]
\vspace{-0.5cm}
 \scalebox{0.68}{
\begin{tabular}{l|ccc|ccc|c|ccc|ccc}
\toprule[0.6mm]
\multirow{2}{*}{Module} & \multicolumn{3}{c|}{WinoGAViL} & \multicolumn{3}{c|}{Winoground} & Whoops     & \multicolumn{3}{c|}{VCR}                                    & \multicolumn{3}{c}{NYCCC}     \\
                        & 5/6      & 10/12     & SWOW    & Text     & Image     & Group    & GPT4 Rate & Q-\textgreater{}A & QA-\textgreater{}R & Q-\textgreater{}AR & Match acc. & CrowdAcc & NYAcc \\ \midrule[0.3mm]
Base                    & 60.0     & 58.3      & 78.4    & 28.7     & 26.2      & 16.0     & 36.4         & 38.0              & 37.0               & 21.3               & 41.8       & 41.3     & 46.0  \\
Base+CaID               & 63.5     & 62.0      & 73.7    & 31.5     & 30.0      & 19.7     & 54.6         & 43.9              & 44.2               & 22.9               & 51.5       & 48.7     & 53.6  \\
Base+CVR-ICL            & 69.8     & 66.1      & 80.9    & 39.0     & 29.2      & 22.0     & 60.6         & 48.8              & 49.2               & 25.8               & 48.0       & 47.6     & 52.9  \\
CVR-LLM$_{GPT4}$                 & \textbf{73.4}     & \textbf{73.2}      & \textbf{86.5}    & \textbf{43.5}     & \textbf{35.0}      & \textbf{26.5}     & \textbf{62.0}         & \textbf{54.3}              & \textbf{52.9}               & \textbf{30.4}               & \textbf{60.6}       & \textbf{57.4}     & \textbf{63.1}  \\
\toprule[0.6mm]
\end{tabular}}
\vspace{-0.2cm}
\caption{The ablation study of our CVR-LLM on five complex visual reasoning tasks. "Base" represents using the general image captions and GPT4 to complete these tasks. "Base+CaID" means using the context-aware image descriptions instead of the general image captions and GPT4 to test the performance. "Base+CVR-ICL" represents using general image captions and GPT4 with our designed CVR-ICL learning methods.}
\vspace{-0.4cm}
\label{tab2}
\end{table*}

\subsection{Analysis}
\begin{figure}[t]
  \centering
   \includegraphics[width=0.85\linewidth]{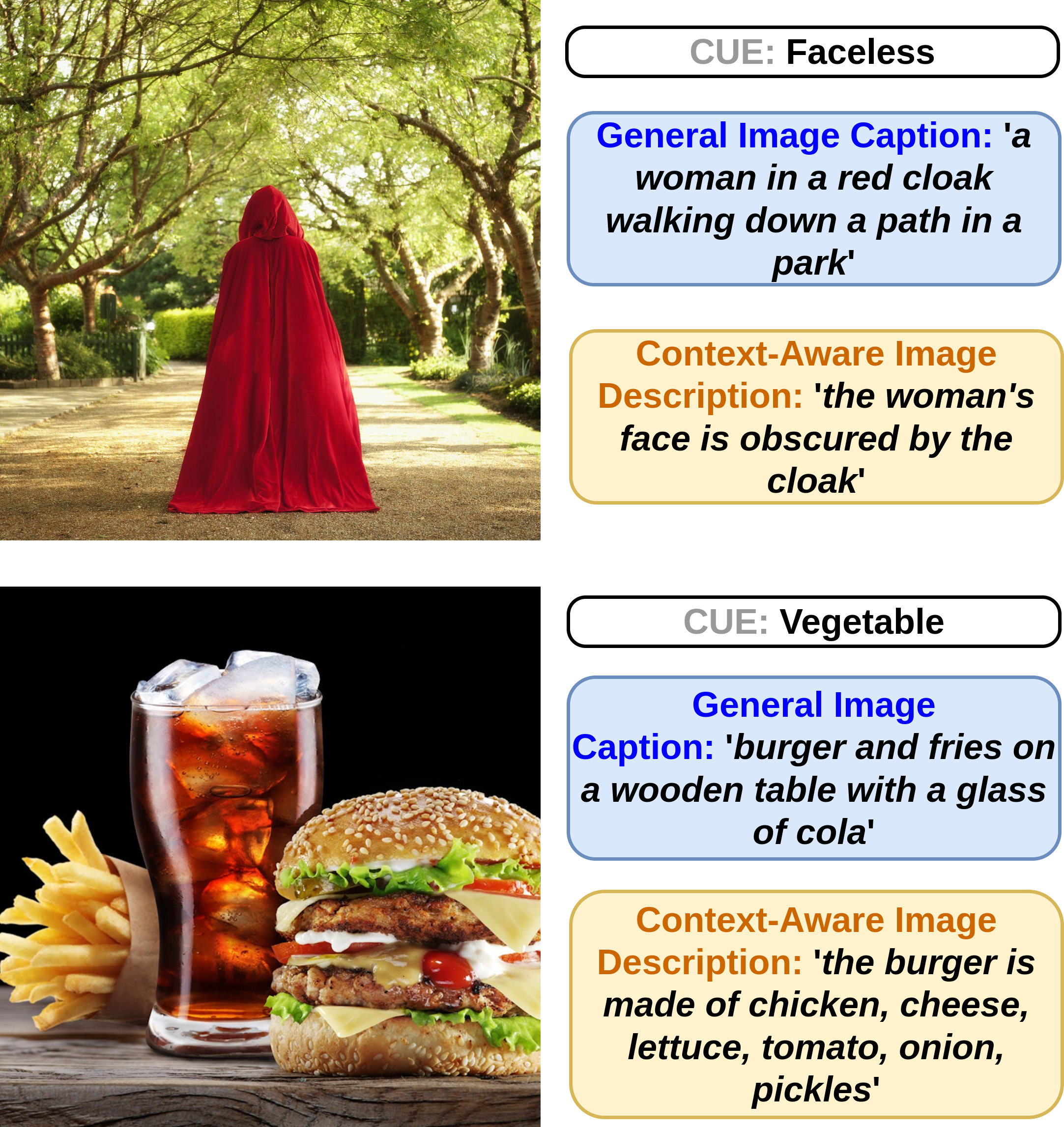}
   \vspace{-0.2cm}
   \caption{Two examples from WinoGAViL compare context-aware image descriptions with general image captions. WinoGAViL is designed to ask the model to select the image that best matches the cue word.}
   \label{fig5}
   \vspace{-0.4cm}
\end{figure}

\begin{figure}[t]
  \centering
   \includegraphics[width=0.9\linewidth]{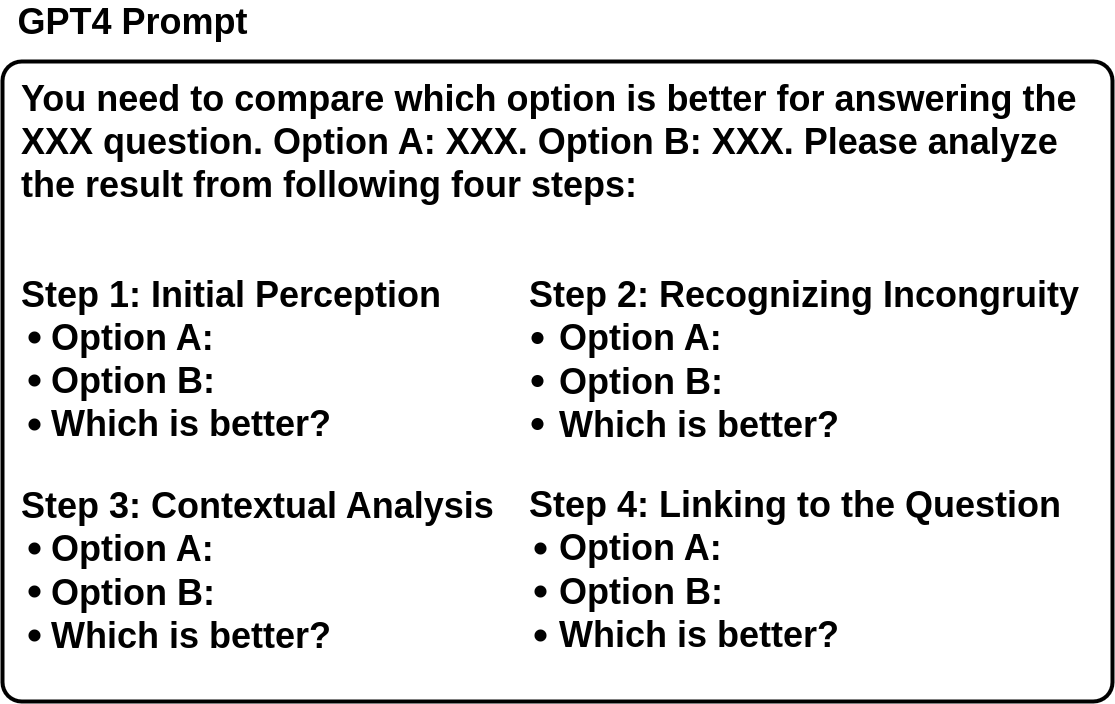}
   \vspace{-0.2cm}
   \caption{The illustration of how to use GPT4 for step-by-step comparsion.}
   \label{fig6}
\end{figure}

\paragraph{Context-Aware Image Description vs General Image Caption}
In this section, we investigate CaID's impact at an abstract level and design a novel method to quantitatively demonstrate the semantic gap between context-aware image descriptions and general image captions (Note that the performance impact has been shown in Table~\ref{tab2}). Figure~\ref{fig5} provides two examples comparing context-aware image descriptions with general image captions and our goal is to determine whether context-aware descriptions offer more contextually relevant information to aid LLMs in decision-making. Unlike traditional sentence evaluations that rely on annotations to compute metrics like BLEU~\cite{papineni2002bleu} and CIDEr~\cite{vedantam2015cider}, we lack direct measures to assess the contextual relevance of sentences. To address this, we use GPT4~\cite{achiam2023gpt} to evaluate the relative effectiveness between two kinds of expressions with the prompt: ``\textit{Evaluate the equivalence of the following two options for the task XXX. Option A: XXX; Option B: XXX. Please return True if Option B is better than Option A in answering questions; return False if the opposite is true; return Equal if they are the same for the question.}''. Additionally, inspired by the concept of chain-of-thought (CoT)~\cite{wei2022chain}, we propose a novel comparison chain-of-comparison (CoC), which implements a step-by-step analysis to evaluate the effectiveness. This method involves a comprehensive four-step analysis protocol, depicted in Figure~\ref{fig6}. It follows a series of cognitive steps that our brains undertake to make sense of information, particularly when engaging with complex problems.

\begin{figure}[t]
  \centering
   \includegraphics[width=0.96\linewidth]{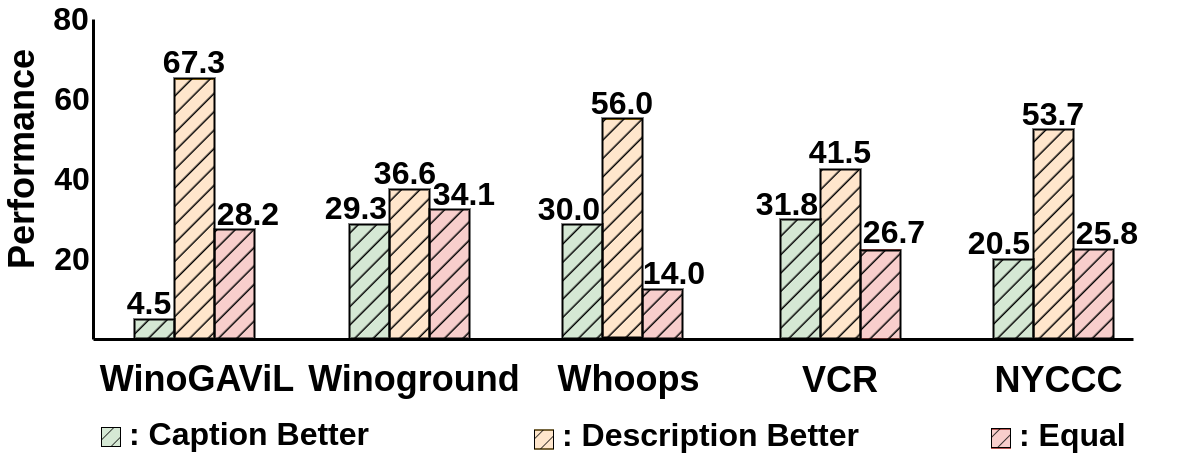}
   \vspace{-0.2cm}
   \caption{Hypothesis verification with GPT4, which demonstrates the effectiveness of our CaID against general image captions.}
   \label{fig7}
   \vspace{-0.5cm}
\end{figure}

\begin{table}[t]
\vspace{-0.7cm}
\scalebox{0.59}{
\begin{tabular}{llccccc}
\toprule[0.7mm]
Dataset                        & Option             & Step 1 & Step 2 & Step 3 & Step 4 & Average \\ \midrule[0.3mm]
\multirow{3}{*}{WinoGAViL}  & Caption Better     & 6.0    & 4.3    & 8.3    & 5.0    & 5.9     \\
                            & Description Better & 75.3   & 76.0   & 71.3   & 76.7   & \textbf{74.8}    \\
                            & Equal              & 18.7   & 19.7   & 20.3   & 18.3   & 19.3    \\ \midrule[0.3mm]
\multirow{3}{*}{Winoground} & Caption Better     & 24.0   & 24.0   & 29.0   & 27.0   & 26      \\
                            & Description Better & 59.0   & 56.0   & 59.0   & 56.0   & \textbf{57.5}    \\
                            & Equal              & 17.0   & 20.0   & 12.0   & 17.0   & 16.5    \\ \midrule[0.3mm]
\multirow{3}{*}{Whoops}     & Caption Better     & 27.0   & 13.0   & 14.0   & 13.0   & 16.7    \\
                            & Description Better & 71.0   & 80.0   & 76.0   & 75.0   & \textbf{75.5}    \\
                            & Equal              & 2.0    & 7.0    & 10.0   & 12.0   & 7.7     \\ \midrule[0.3mm]
\multirow{3}{*}{VCR}     & Caption Better     & 24.3   & 32.5   & 30.1   & 28.6   & 28.9    \\
                            & Description Better & 53.5   & 45.4   & 50.6   & 52.7   & \textbf{50.5}    \\
                            & Equal              & 22.2    &  22.1   & 19.3   & 18.7   & 20.6     \\ \midrule[0.3mm]
\multirow{3}{*}{NYCCC}     & Caption Better     & 18.6   & 15.8   & 17.4   & 19.1   & 17.7    \\
                            & Description Better & 58.5   & 62.3   & 60.4   & 61.0   & \textbf{60.5}    \\
                            & Equal              & 22.9    & 21.9    & 22.2   & 19.9   & 21.8     \\ \bottomrule[0.7mm]
\end{tabular}}
\vspace{-0.2cm}
\caption{The performance of using GPT4 to assess the effectiveness of two options (general image caption and our context-aware image description) based on CoC.}
\label{tab3}
\vspace{-0.1cm}
\end{table}

Figure~\ref{fig7} shows the results of directly employing GPT4 to compare the effectiveness of general image captions with our image descriptions in the specific scenario of answering task-related questions. Furthermore, Table~\ref{tab3} presents the performance derived from utilizing GPT4 to conduct a detailed, step-by-step analytical assessment of effectiveness. These empirical results indicate that our approach yields image descriptions with enhanced contextual relevance, thereby significantly aiding LLMs in the decision-making process, particularly on the WinoGAViL and Whoops datasets.

\paragraph{Complex Visual Reasoning ICL vs Other ICL}
The CVR-ICL is designed to optimize the selection of in-context exemplars within a multi-modal environment, thereby enhancing the reasoning abilities of LLMs. This innovative method is contrasted with three alternative configurations: Random In-Context Learning (RICL)~\cite{brown2020language}, KATE~\cite{liu2021makes}, and Multi-modal Similar In-Context Learning (MMICL)~\cite{zhao2023mmicl}. To ensure a fair comparison, we utilized general image captions across all models to test performance for eliminating the effect of our context-aware image descriptions. As demonstrated in Table~\ref{tab4}, our CVR-ICL outperforms other ICL methods, demonstrating its adeptness at integrating and leveraging both textual and multi-modal domains to select the most contextually appropriate exemplars.

\begin{table}[t]
\scalebox{0.54}{
\begin{tabular}{llcccc}
\toprule[0.5mm]
Dataset                        & Category      & RICL~\shortcite{brown2020language} & KATE~\shortcite{liu2021makes} & MMICL~\shortcite{zhao2023mmicl} & CVR-ICL \\ \midrule[0.3mm]
\multirow{3}{*}{WinoGAViL}  & 5/6         & 64.1  & 68.6   & 66.3   & \textbf{69.8}   \\
                            & 10/12       & 61.7  & 64.1   & 62.8   & \textbf{66.1}   \\
                            & SWOW        & 80.7  & \textbf{82.8}     & 80.9   & 80.9   \\ \midrule[0.3mm]
\multirow{3}{*}{Winoground} & Text  & 35.0  & 29.5  & 27.5   & \textbf{39.0}   \\
                            & Image  & 22.5  & 30.0   & 25.0   & \textbf{29.2}   \\
                            & Group & 18.5  & \textbf{20.0}   & 17.5   & 22.0   \\ \midrule[0.3mm]
Whoops                      & GPT4 Rate & 60.4  & 62.0   & 60.8   & \textbf{62.0}  \\ \midrule[0.3mm]
\multirow{3}{*}{VCR}        & Q->A  & 45.1  & 48.6  & 44.0   & \textbf{48.8}   \\
                            & QA->R  & 46.5  & 48.9   & 46.3   & \textbf{49.2}   \\
                            & Q->AR & 22.5  & 24.8   & 23.6   & \textbf{25.8}   \\ \midrule[0.3mm]
\multirow{3}{*}{NYCCC}      & Match acc.  & 44.4  & 47.5  & 45.5   & \textbf{48.0}   \\
                            & CrowdAcc  & 46.6  & 46.4   & 43.7   & \textbf{47.6}   \\
                            & NYAcc &50.3   & 51.2   & 49.8   & \textbf{52.9}   \\ 
\bottomrule[0.5mm]
\end{tabular}}
\vspace{-0.1cm}
\caption{The performance of using different ICL methods on different datasets.}
\label{tab4}
\vspace{-0.4cm}
\end{table}

\begin{figure}[t]
\vspace{-0.8cm}
  \centering
   \includegraphics[width=0.68\linewidth]{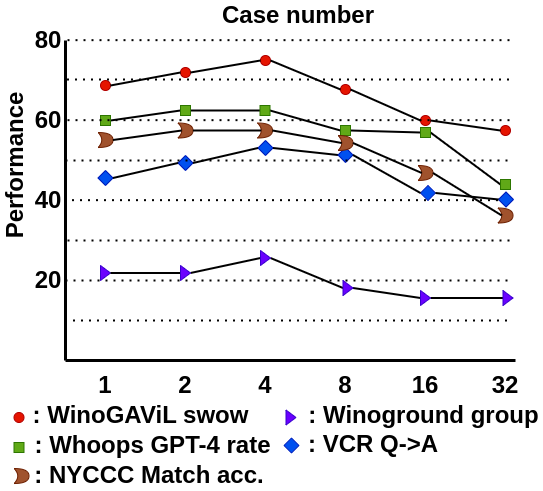}
   \vspace{-0.2cm}
   \caption{The different case numbers in CVR-ICL and corresponding performance.}
   \label{fig8}
   \vspace{-0.3cm}
\end{figure}

\paragraph{Case Number Selection in Complex Visual Reasoning ICL}
Figure~\ref{fig8} illustrates the influence of varying case numbers in the CVR-ICL on the performance of our proposed CVR-LLM method. The experimental results suggest a trend where the model's performance initially improves with an increase in case numbers, exhibits fluctuations at higher numbers, and eventually declines as the case number becomes excessively large. This pattern suggests that the optimal selection for the number of cases is four.

\begin{figure}[t]
  \centering
   \includegraphics[width=0.88\linewidth]{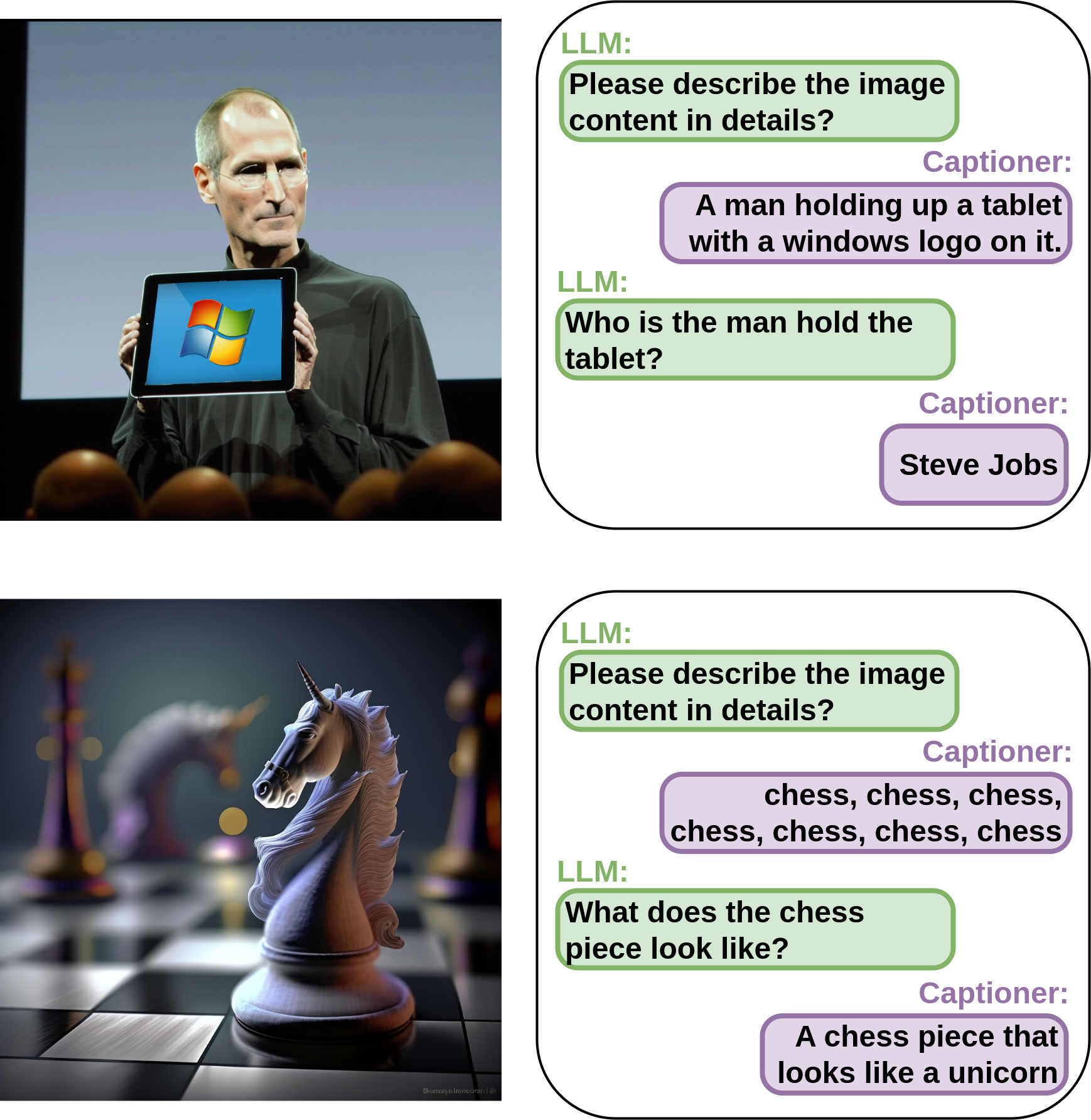}
   \vspace{-0.2cm}
   \caption{Two qualitative results from Whoops illustrating the capabilities of our approach. Whoops is designed to ask the model to explain what makes images weird.}
   \label{fig9}
   \vspace{-0.4cm}
\end{figure}

\section{Qualitative Results}
\label{qualitative}
To showcase the capabilities of our approach, we present qualitative results in Figure~\ref{fig9}. It illustrates how LLMs leverage contextual information to ask more relevant and insightful questions tailored the specific tasks. For instance, when provided with an image of the chess piece, the LLMs might ask ``What does the chess piece look like?''. Subsequently, the captioner model generates contextually appropriate descriptions, such as ``A chess piece that looks like a unicorn.''. This synergy enhances the LLM's decision-making process, making it more precise and context-aware. More detailed qualitative results with corresponding prompts and CVR-ICL examples are illustrated in Appendix~\ref{A1} and Appendix~\ref{A2}.

\section{Conclusion}
In this work, we propose CVR-LLM, an innovative approach for complex visual reasoning tasks. This method boosts LLMs' understanding of visual content for complex reasoning via context-aware image descriptions. We also develop a multi-modal in-context learning technique, enhancing LLMs' reasoning skills at both image and text levels. Experimental results show that CVR-LLM sets new benchmarks across multiple complex visual reasoning tasks. We also introduce a nuanced GPT4 based analysis technique Chain-of-Comparison to automatically break down and contrast among various aspects of generated results. 

\section{Limitation}
\begin{figure}[ht]
  \centering
   \includegraphics[width=0.9\linewidth]{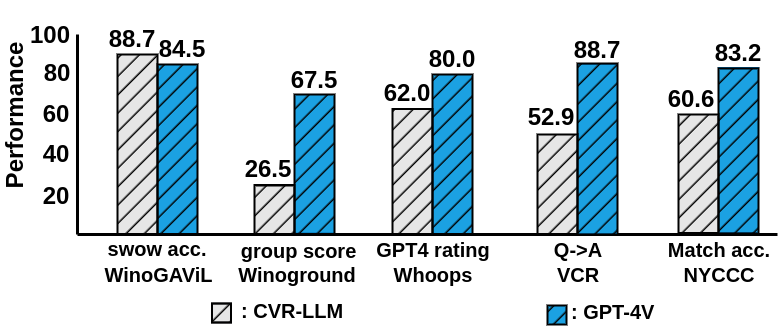}
   \vspace{-0.2cm}
   \caption{The comparison of our CVR-LLM against GPT-4V.}
   \label{fig10}
   \vspace{-0.4cm}
\end{figure}

Although our approach achieves SOTA performance across a wide range of complex visual reasoning benchmarks, it still has two notable limitations. First, compared to the MLLMs that can perform end-to-end inference directly, our approach operates as an LLM-agent-driven framework. This involves VLMs generating context-aware image descriptions, followed by the LLM performing inference with ICL to predict the answer. While this two-step process enhances contextual understanding and reasoning, it may significantly increase time consumption compared to direct end-to-end inference models. Second, despite its overall strong performance and generalization ability, our approach still lags behind GPT4V in some tasks. Figure~\ref{fig10} shows that our CVR-LLM can surpass GPT4V in SWOW setting in WinoGAViL dataset but fall short in others. Our future work will focus on refining the integration between VLMs and LLMs components and enhancing the model’s efficiency and accuracy across a broader spectrum of complex visual reasoning challenges.

\bibliography{custom}

\clearpage
\appendix
\section{Appendix}
\subsection{Qualitative Results with Corresponding Prompt}
\label{A1}
\begin{figure}[ht]
  \centering
   \includegraphics[width=0.95\linewidth]{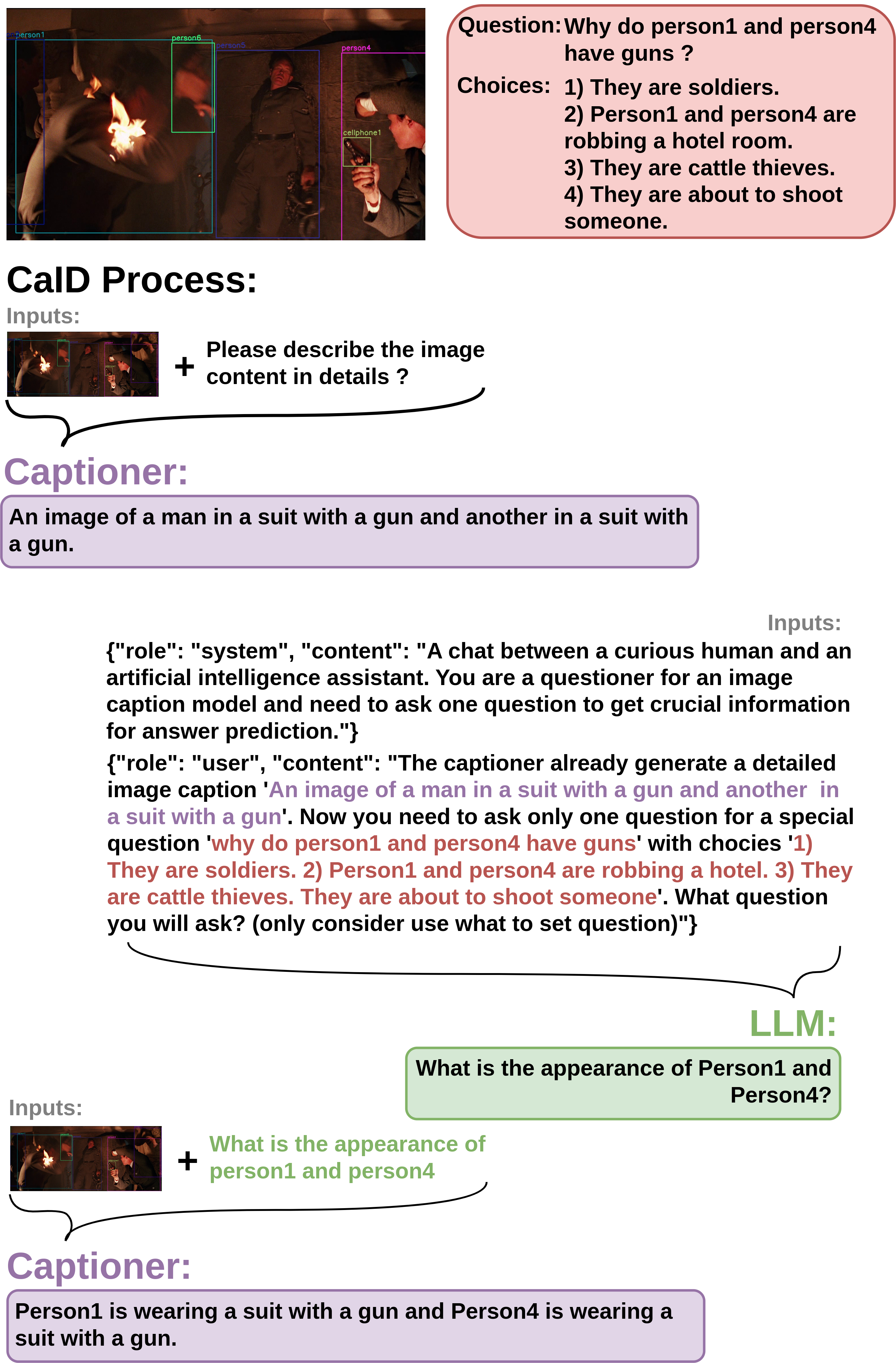}
   \vspace{-0.1cm}
   \caption{The detailed illustration of our CaID process on VCR. Best viewed by zooming in.}
   \label{fig11}
   \vspace{-0.2cm}
\end{figure}

Section~\ref{qualitative} only illustrates the simplified process of our Context-aware Image Description (CaID) generation. Here, we delve into more details about the generation process and the corresponding prompts. Figure~\ref{fig11} provides an example of the CaID generation process applied to the VCR~\cite{zellers2019recognition} task. In this example, the initial input consists of an image showing several individuals, with two of them (Person1 and Person4) holding guns. The associated question is: ``Why do Person1 and Person4 have guns?'' with multiple-choice options such as ``1) They are soldiers. 2) Person1 and Person4 are robbing a hotel room. 3) They are cattle thieves. 4) They are about to shoot someone.''. 

The CaID process begins by generating a detailed description of the image. The captioner model produces an initial caption: ``An image of a man in a suit with a gun and another in a suit with a gun.''. This caption, while descriptive, lacks the context needed to answer the specific question posed. To address this, our system prompts the LLM with a scenario where it acts as a questioner for the image caption model. The LLM is instructed to generate a follow-up question to gather crucial information for answer prediction. The prompt guides the LLM to consider specific details such as the appearance and pose of the individuals. In this case, the LLM generates the question: ``What is the appearance of Person1 and Person4?''. This question is designed to extract more contextually relevant details from the image captioner. The captioner then provides a refined description: ``Person1 is wearing a suit with a gun and Person4 is wearing a suit with a gun.''. This additional information helps to better understand the scene and narrows down the possible answers to the original question. This detailed process highlights how our system leverages both multi-modal and textual information to generate precise and contextually relevant descriptions, ultimately improving the performance on complex visual reasoning tasks.

\begin{figure*}[t]
  \centering
   \includegraphics[width=0.95\linewidth]{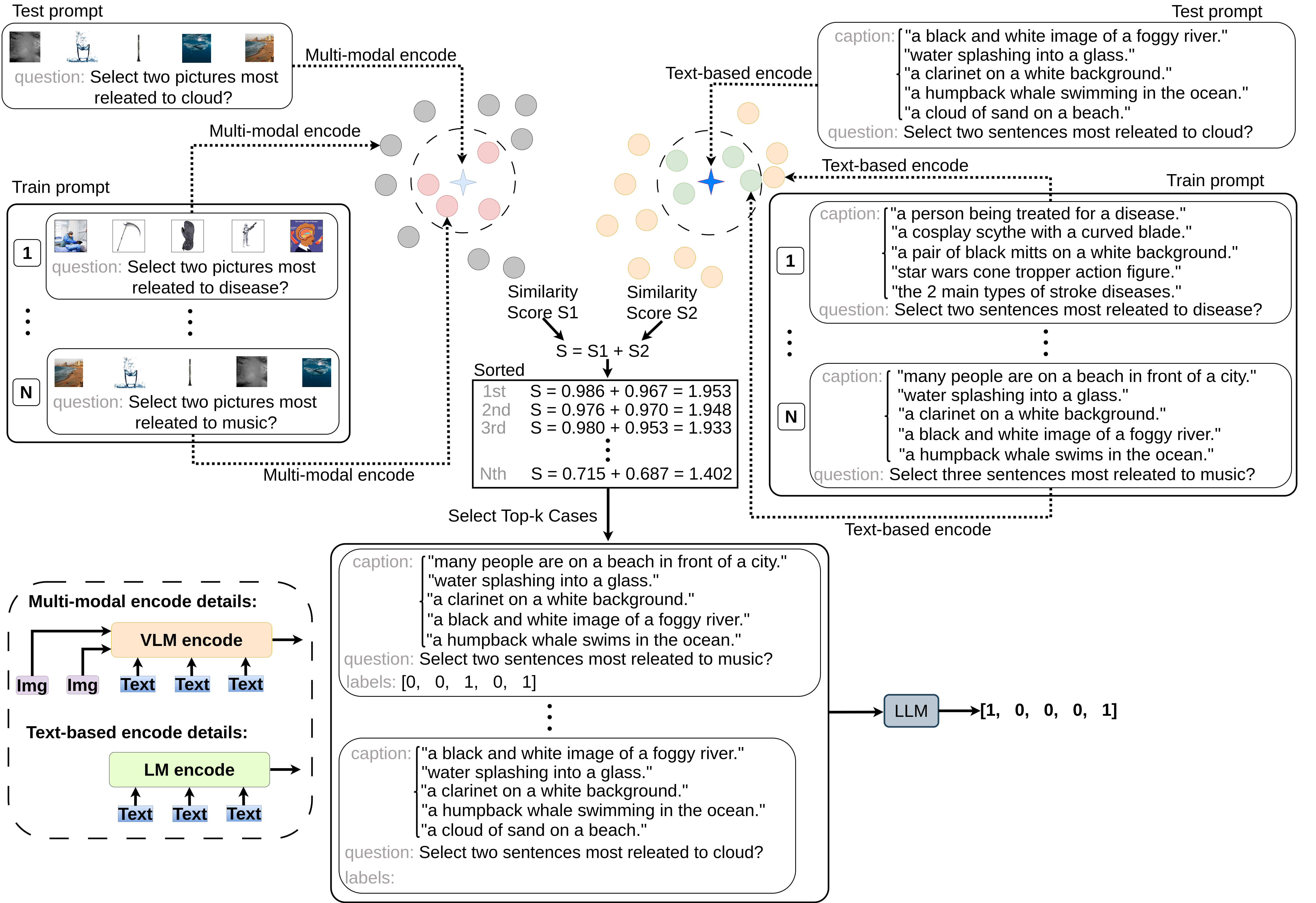}
   \vspace{-0.0cm}
   \caption{The detailed illustration of our CVR-ICL on WinoGAViL. Best viewed by zooming in.}
   \label{fig13}
   \vspace{-0.2cm}
\end{figure*}

\subsection{Qualitative CVR-ICL Examples}
\label{A2}
Section~\ref{in-context} only illustrates the mechanism of our CVR-ICL. Here, we explain more details about its implementation. Figures~\ref{fig13}  showcases one example of our CVR-ICL on the WinoGAViL~\cite{bitton2022winogavil}.

To accurately calculate similarity scores using the cosine similarity function, we utilize BM25~\cite{robertson1995okapi} for text encoding and BLIP2 multi-embedding~\cite{li2023blip} for multi-modal inputs. As illustrated in Figure~\ref{fig13}, the process begins with encoding both the test and training prompts through multi-modal and text-based encoders. For instance, a test case from WinoGAViL might contain the question ``Select two pictures most related to clouds?'' along with images of a foggy river, a cloud of sand on a beach, and other related scenes. At the beginning, the multi-modal encoder processes these images as well as the question and generates multimodal-level embeddings. Simultaneously, we convert these images into context-aware image descriptions and translate the entire case into text form. The text-based encoder then generates corresponding text-level embeddings. Next, we calculate the individual cosine similarity scores in both the multi-modal and text domains. The final similarity score, which determines the most relevant cases, is calculated in a balanced manner as $S = S_1 + S_2$. These scores are then sorted, and the top-$k$ most similar cases are selected as in-context learning examples. This dual-encoding and similarity scoring approach ensures that we capture the nuanced relationships between multi-modal inputs and text, thereby enhancing the accuracy and relevance of our in-context learning framework.

\subsection{Comparative Analysis with Fine-tuned Models}
In this section, we explore the impact of fine-tuning strategy on performance in complex visual reasoning tasks. Since some tasks in the complex visual reasoning field are initially designed in the supervised setting, we are curious whether our approach can also perform better with the help of real annotation. For the test-only datasets WinoGAViL and Winoground, we randomly divided them into splits of $80\%$ training, $10\%$ validation, and $10\%$ testing.  Due to the small number of cases in these tasks, we abandoned training LLMs to avoid catastrophic forgetting. Instead, we chose to fine-tune the captioner using the real labels and incorporated these real annotations into our CVR-ICL examples. Results shown in Table~\ref{tab6} compare our CVR-LLM's performance in zero-shot and fine-tuned settings against SOTA performances, revealing that our method maintains SOTA performance in several areas.

\begin{table}[ht]
\scalebox{0.65}{
\begin{tabular}{llcccc}
\toprule[0.5mm]
\multirow{2}{*}{Dataset}    & \multirow{2}{*}{Category} & \multicolumn{2}{c}{Zero-shot} & \multicolumn{2}{c}{Finetuned} \\
                            &                           & SOTA         & CVR-LLM        & SOTA         & CVR-LLM        \\ \midrule[0.3mm]
\multirow{3}{*}{WinoGAViL} & 5/6                       & 55.0         & 74.7           & 54.6         & \textbf{82.8}           \\
                            & 10/12                     & 52.0         & 73.2           & 47.2         & \textbf{80.8}           \\
                            & SWOW                      & 59.0         & 88.7           & 68.8         & \textbf{95.9}           \\ \midrule[0.3mm]
\multirow{3}{*}{Winoground} & Text                      & 46.5         & 43.5           & 47.0         & \textbf{55.0}           \\
                            & Image                     & 27.7         & 35.0           & 42.2         & \textbf{42.5}           \\
                            & Group                     & 24.2         & 26.5           & 30.5         & \textbf{35.0}           \\ \midrule[0.3mm]
Whoops                      & GPT-4 Rate                & 31.0         & 62.0           & 71.0         & \textbf{72.0}           \\ \midrule[0.3mm]
\multirow{3}{*}{VCR}        & Q-\textgreater{}A         & 48.8         & 52.9           & \textbf{87.4}         & 85.3           \\
                            & QA-\textgreater{}R        & 49.7         & 54.3           & \textbf{89.6}         & 87.5           \\
                            & Q-\textgreater{}AR        & 28.6         & 30.4           & \textbf{78.6}         & 77.1           \\ \midrule[0.3mm]
\multirow{3}{*}{NYCCC}      & Match acc.                & 60.4         & 60.6           & \textbf{84.5}         & 80.9           \\
                            & CrowdAcc                  & 59.2         & 57.4           & \textbf{73.3}         & 69.6           \\
                            & NYAcc                     & 66.5         & 63.1           & \textbf{68.2}         & 65.4           \\ \bottomrule[0.5mm]
\end{tabular}}
\caption{The comparison of our CVR-LLM against SOTA performance under two kinds of settings.}
\label{tab6}
\vspace{-0.2cm}
\end{table}

\begin{figure*}[t]
  \centering
   \includegraphics[width=0.7\linewidth]{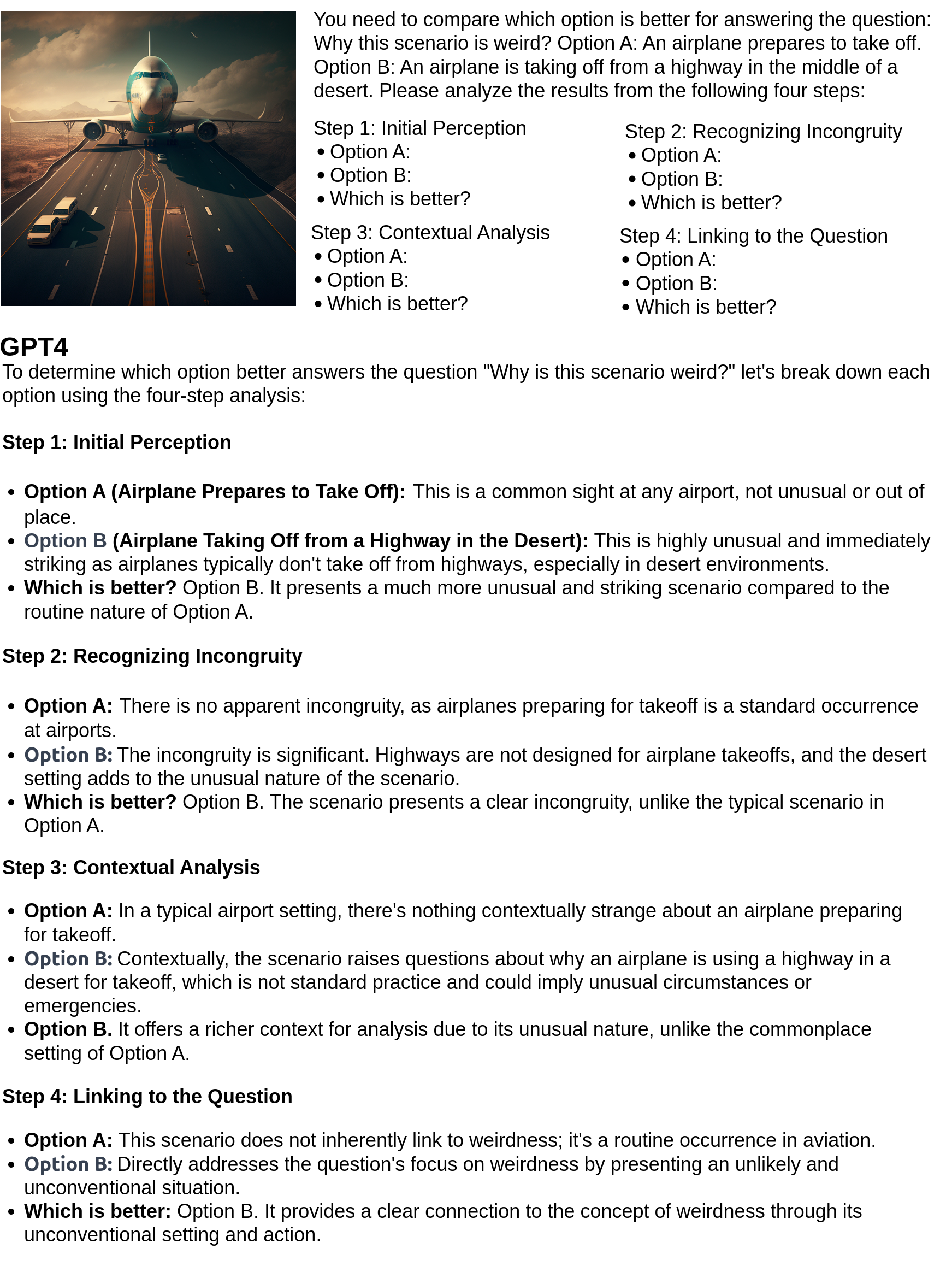}
   \vspace{-0.1cm}
   \caption{The detailed illustration of our CoC on Whoops. Best viewed by zooming in.}
   \label{fig15}
   \vspace{-0.1cm}
\end{figure*}

\subsection{More Explanation about Our CoC}
The Chain-of-Comparison (CoC) is designed to qualitatively analyze the semantic contribution of context-aware image descriptions against general image captions. It is inspired by the popular idea of Chain-of-Thought, which implements a step-by-step analysis to evaluate effectiveness. Figure~\ref{fig15} shows an example from the Whoops dataset, comparing the semantic gap between a general caption ``An airplane prepares to take off'' (Option A) and our context-aware image description ``An airplane is taking off from a highway in the middle of the desert.''. (Option B).

Our CoC prompt asks the LLM to analyze the semantic contribution through four steps: Initial Perception, Recognizing Incongruity, Contextual Analysis, and Linking to the Question. This process mimics the human brain's analytical process. We directly ask the LLM to compare the contributions of the two options and determine which is better. 

For instance, in the Initial Perception step, the LLM identifies Option B as superior because it is highly unusual and immediately striking, as airplanes typically do not take off from highways, especially in desert environments. This scenario is much more unusual and striking compared to the routine scenario of Option A, which merely depicts an airplane preparing to take off at an airport. During the Contextual Analysis step, Option B is again favored. The LLM explains that contextually, the scenario raises questions about why an airplane is using a highway in a desert for takeoff, which is not standard practice and could imply unusual circumstances or emergencies. Option A, in contrast, has nothing contextually strange about an airplane preparing for takeoff in a typical airport setting. Finally, in the Linking to the Question step, the LLM determines that Option B provides a clearer connection to the concept of weirdness through its unconventional and striking situation. Option A does not inherently link to weirdness, as it describes a routine occurrence in aviation.

This example demonstrates how our CoC framework effectively breaks down and evaluates the semantic contributions of different types of image descriptions, highlighting the advantages of context-aware image descriptions in complex visual reasoning tasks.

\subsection{The CVR-LLM Performance with LLaMA2}
Table~\ref{tab1} presents the results of our CVR-LLM framework using Llama3, GPT-3.5, and GPT-4 base models. Additionally, we evaluated CVR-LLM on the Llama2-13B model~\cite{touvron2023llama2}, which was also employed in LLaVA~\cite{wu2023visual, liu2024visual}, to ensure a fair comparison. Table~\ref{tab7} compares the performance of CVR-LLM (Llama2-based) and CVR-LLM (Llama3-based) against LLaVA versions 1.0~\cite{liu2024visual} and 1.5~\cite{wu2023visual} on complex reasoning tasks. The results demonstrate that while our CVR-LLM performs well on the Llama2 base model, it slightly underperform compared to Llama3.
\begin{table}[t]
\scalebox{0.73}{
\begin{tabular}{lccc}
\toprule[0.5mm]
Model             & Whoops & VCR (Q-\textgreater{}A) & NYCCC (Match) \\ \hline
LLaVA 1.0         & 32.0   & 28.3                    & 55.8          \\
LLaVA 1.5         & 42.4   & 35.1                    & 59.3          \\
CVR-LLM$_{Llama2}$ & 55.6   & 44.6                    & 56.4          \\
CVR-LLM$_{Llama3}$ & 60.4   & 50.5                    & 59.8          \\\bottomrule[0.5mm]
\end{tabular}}
\caption{The comparison of our CVR-LLM with Llama2 and Llama3 base against SOTA LLaVA models.}
\label{tab7}
\end{table}

\begin{table}[t]
\scalebox{0.76}{
\begin{tabular}{lcccccc}
\toprule[0.5mm]
$\alpha$                & 0.1  & 0.2  & 0.3  & 0.5  & 1    & 2    \\ \hline
WinoGAViL (5/6)   & 66.6 & 67.9 & 66.5 & 68.3 & 69.8 & 65.8 \\
WinoGAViL (10/12) & 63.7 & 65.1 & 63.6 & 64.8 & 66.1 & 62.0 \\
WinoGAViL (swow)  & 76.3 & 77.0 & 75.8 & 78.1 & 80.9 & 72.7 \\\bottomrule[0.5mm]
\end{tabular}}
\caption{The performance of our CVR-LLM framework with varying $\alpha$ values on the WinoGAViL dataset.}
\label{tab8}
\end{table}

\subsection{The Parameter Setting in Equation 4c}
Section~\ref{in-context} explains that our in-context learning examples are selected based on a similarity score calculated as follows: 
\begin{subequations} 
\begin{align} 
&s = \alpha * s_m + s_t, \quad (\alpha = 1). \label{eq-5} \end{align} 
\end{subequations} 
In this section, we discuss how the parameter $\alpha$ influences the performance of In-Context Learning (ICL). Table~\ref{tab8} presents the results for various values of $\alpha$ on the WinoGAViL dataset. The results indicate that $\alpha = 1$ leads to the best performance of our CVR-ICL strategy.

\subsection{Comparison against Other VLM+LLM Methods}
In the main paper, we compare our method with several popular end-to-end MLLMs, including LLaVA~\cite{wu2023visual} and MiniGPT-4~\cite{zhu2023minigpt}. Additionally, we evaluate our approach against VLM+LLM methods such as DDCoT~\cite{zheng2023ddcot} and DIEM~\cite{jiang2024diem}. Table~\ref{tab9} presents the comparison results of our CVR-LLM framework versus these methods. While our approach is similar to DIEM in focusing on visual information from images, it demonstrates superior performance in complex visual reasoning tasks. Instead of decomposing the image and extracting information from individual components, we utilize an iterative refinement strategy, enabling the Large Language Model (LLM) to pose more precise questions and extract highly specific, valuable information from the image.

\begin{table}[t]
\scalebox{0.7}{
\begin{tabular}{lccccc}
\toprule[0.5mm]
\multirow{2}{*}{Models} & \multirow{2}{*}{\begin{tabular}[c]{@{}c@{}}WinoGAViL\\ (swow)\end{tabular}} & \multirow{2}{*}{\begin{tabular}[c]{@{}c@{}}Winoground\\ (group)\end{tabular}} & \multirow{2}{*}{\begin{tabular}[c]{@{}c@{}}Whoops\\ (GPT4 reate)\end{tabular}} & \multirow{2}{*}{\begin{tabular}[c]{@{}c@{}}VCR\\ (Q-\textgreater{}A)\end{tabular}} \\
                        &                                                                             &                                                                               &                                                                                &                                                                                    \\ \hline
DDCoT                   & 77.5                                                                        & 20.2                                                                          & 48.4                                                                           & 40.7                                                                               \\
DIEM                    & 83.5                                                                        & 22.5                                                                          & 58.0                                                                           & 50.5                                                                               \\
CVR-LLM                 & 86.5                                                                        & 26.5                                                                          & 62.0                                                                           & 54.3                                                                               \\ \bottomrule[0.5mm]
\end{tabular}}
\caption{The comparison of our CVR-LLM against other VLM+LLM methods.}
\label{tab9}
\end{table}

\subsection{The Performance on Multi-step Reasoning Dataset}
Our CVR-LLM framework is designed for complex visual reasoning tasks, making it well-suited for multi-step reasoning datasets, such as ScienceQA~\cite{lu2022learn} and M3CoT~\cite{chen2024m}. In this section, we evaluate the performance of our CVR-LLM on the M3CoT dataset to determine its effectiveness. Table~\ref{tab10} presents a comparison between our CVR-LLM and other Tool-Usage methods. The results show that our approach performs well on questions related to general image content, particularly in areas like physical and social sciences. However, it faces challenges with images containing multiple elements, occasionally leading to hallucinations in detailed descriptions.
\begin{table}[t]
\scalebox{0.72}{
\begin{tabular}{lccccc}
\toprule[0.5mm]
Model    & HuggingGPT & IdealGPT & Chameleon & CVR-LLM \\ \hline
Lang     & 17.57      & 31.73    & 43.87     & 34.10   \\
Natural  & 20.93      & 31.63    & 26.05     & 33.20   \\
Social   & 10.33      & 26.23    & 25.44     & 24.84   \\ \hline
Physical & 8.7        & 56.52    & 39.13     & 71.11   \\
Social   & 14.75      & 50.00    & 37.30     & 69.83   \\
Temporal & 9.76       & 26.83    & 48.78     & 30.89   \\ \hline
Algebra  & 11.35      & 20.57    & 17.73     & 29.29   \\
Geometry & 22.50      & 30.00    & 26.25     & 22.50   \\
Theory   & 9.52       & 38.10    & 23.81     & 28.57   \\ \bottomrule[0.5mm]
\end{tabular}}
\caption{The comparison of our CVR-LLM against other Tool-Usage methods on the M3CoT dataset.}
\label{tab10}
\end{table}

\end{document}